\documentclass[11pt,a4paper]{article}
\usepackage[hyperref]{acl2019}
\usepackage{times}
\usepackage{latexsym}
\usepackage{url}
\usepackage{amsmath}
\usepackage{multirow}
\usepackage{bbm}
\usepackage{graphicx}
\usepackage{amssymb}
\usepackage{booktabs}
\usepackage{adjustbox}
\usepackage{xspace}

\aclfinalcopy 

\newcommand{\secref}[2][]{Section#1~\ref{sec:#2}}

\newcommand{\tabref}[2][]{Table#1~\ref{tab:#2}}
\newcommand{\figref}[2][]{Figure#1~\ref{fig:#2}}

\newcommand{\eqnref}[2][]{Equation#1~(\ref{eqn:#2})}

\newcommand{\anoteone}[2][]{\makebox[0pt][l]{$^{(1)}$}}
\newcommand{\anotetwo}[2][]{\makebox[0pt][l]{$^{(2)}$}}

\newcommand{\rstrel}[1]{\texttt{#1}\xspace}
\DeclareMathOperator{\softmax}{softmax}

\title{Improved Document Modelling with a Neural Discourse Parser}

\author{Fajri Koto \qquad Jey Han Lau \qquad Timothy Baldwin\\
  School of Computing and Information Systems \\
  The University of Melbourne \\
  \texttt{\small ffajri@student.unimelb.edu.au, jeyhan.lau@gmail.com, 
 tbaldwin@unimelb.edu.au} \\
 }

\begin{document}	\maketitle
    \begin{abstract}
        Despite the success of attention-based neural models for natural 
        language generation and classification tasks, they are unable to 
capture the discourse structure of larger documents. We hypothesize that 
explicit discourse representations have utility for NLP tasks over 
longer documents or document sequences, which sequence-to-sequence 
models are unable to capture. For abstractive summarization, for 
instance, conventional neural models simply match source documents and 
the summary in a latent space without explicit representation of text 
structure or relations. In this paper, we propose to use neural 
discourse representations obtained from a rhetorical structure theory 
(RST) parser to enhance document representations. Specifically, document 
representations are generated for discourse spans, known as the 
elementary discourse units (EDUs). We empirically investigate the 
benefit of the proposed approach on two different tasks: abstractive 
summarization and popularity prediction of online petitions. We find 
that the proposed approach leads to improvements in all cases.
		
	\end{abstract}
	
	\section{Introduction}
	
	Natural language generation and document classification have been widely conducted using neural sequence models based on the encoder--decoder architecture. The underlying technique relies on the production of a context vector as the document representation, to estimate both tokens in natural language generation and labels in classification tasks. By combining recurrent neural networks with attention \cite{bahdanau2015neural}, the model is able to learn contextualized representations of words at the sentence level. However, higher-level concepts, such as discourse structure beyond the sentence, are hard for an RNN to learn, especially for longer documents. We hypothesize that NLP tasks such as summarization and document classification can be improved through the incorporation of discourse information.
	
	In this paper, we propose to incorporate latent representations of discourse units into neural training. A discourse parser can provide information about the document structure as well as the relationships between discourse units. In a summarization scenario, for example, this information may help to remove redundant information or discourse disfluencies. In the case of document classification, the structure of the text can provide valuable hints about the document category. For instance, a scientific paper follows a particular discourse narrative pattern, different from a short story. Similarly, we may be able to predict the societal influence of a document such as a petition document, in part, from its discourse structure and coherence.
	
	Specifically, discourse analysis aims to identify the organization of a text by segmenting sentences into units with relations. One popular representation is Rhetorical Structure Theory (RST) proposed by \newcite{mann1988rhet}, where the document is parsed into a hierarchical tree, where leaf nodes are the segmented units, known as Entity Discourse Units (EDUs), and non-terminal nodes define the relations. 
	
	\begin{figure}
		\centering
		\includegraphics[width=3in]{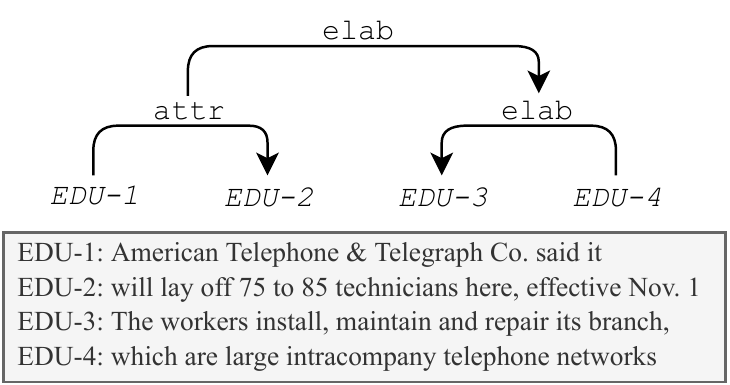}
		\caption{An example of a discourse tree, from \cite{yu2018transition}; \rstrel{elab} = elaboration; \rstrel{attr} = attribute.}
		\label{fig:rsttree}
	\end{figure}
	
    As an example, in \figref{rsttree} the two-sentence text has been 
    annotated with discourse structure based on RST, in the form of 4 
EDUs connected with discourse labels 
\rstrel{attr} and \rstrel{elab}. Arrows in the tree capture the 
nuclearity of relations, wherein a ``satellite'' points to its 
``nucleus''. The \textit{Nucleus} unit is considered more prominent than 
the \textit{Satellite}, indicating that the \textit{Satellite} is a 
supporting sentence for the \textit{Nucleus}. Nuclearity relationships 
between two EDUs can take the following three forms: 
{Nucleus--Satellite}, {Satellite--Nucleus}, and {Nucleus--Nucleus}. In 
this work, we use our reimplementation of the state of the art neural 
RST parser of \citet{yu2018transition}, which is based on eighteen 
relations: \rstrel{purp, cont, attr, evid, comp, list, back, same, 
topic, mann, summ, cond, temp, eval, text, cause, prob, 
elab}.\footnote{The details of each relation can be found on the RST 
website \url{http://www.sfu.ca/rst/index.html}} 
	
    This research investigates the impact of discourse representations 
    obtained from an RST parser on natural language generation and 
document classification. We primarily experiment with an abstractive 
summarization model in the form of a pointer--generator network 
\cite{see2017get}, focusing on two factors: (1) whether summarization 
benefits from discourse parsing; and (2) how a pointer--generator 
network guides the summarization model when discourse information is 
provided. For document classification, we investigate the content-based 
popularity prediction of online petitions with a deep regression model 
\cite{subramanian2018content}. We argue that document structure is a key 
predictor of the societal influence (as measured by signatures to the 
petition) of a document such as a petition.
	
	Our primary contributions are as follows: (1) we are the first to incorporate a neural discourse parser in sequence training; (2) we empirically demonstrate that a latent representation of discourse structure enhances the summaries generated by an abstractive summarizer; and (3) we show that discourse structure is an essential factor in modelling the popularity of online petitions. 
	
	\section{Related Work}
	
	Discourse parsing, especially in the form of RST parsing, has been the target of research over a long period of time, including pre-neural feature engineering approaches \cite{hernault2010hilda,feng2012text,ji2014rep}. Two approaches have been proposed to construct discourse parses: (1) bottom-up construction, where EDU merge operations are applied to single units; and (2) transition parser approaches, where the discourse tree is constructed as a sequence of parser actions. Neural sequence models have also been proposed. In early work,  \citet{li2016discourse} applied attention in an encoder--decoder framework and slightly improved on a classical feature-engineering approach. The current state of the art is a neural transition-based discourse parser \cite{yu2018transition} which incorporates implicit syntax features obtained from a bi-affine dependency parser \cite{dozat2017deep}. In this work, we employ this discourse parser to generate discourse representations.
	
	\subsection{Discourse and Summarization}
	
	Research has shown that discourse parsing is valuable for summarization. Via the RST tree, the salience of a given text can be determined from the nuclearity structure. In extractive summarization, \citet{ono1994abstract}, \citet{donnel1997var}, and \citet{marcu997from} suggest introducing penalty scores for each EDU based on the nucleus--satellite structure. In recent work, \citet{schrimpf2018using} utilizes the \rstrel{topic} relation to divide documents into sentences with similar topics. Every chunk of sentences is then summarized in extractive fashion, resulting in a concise summary that covers all of the topics discussed in the passage. 
	
    Although the idea of using discourse information in summarization is 
    not new, most work to date has focused on extractive summarization, 
where our focus is abstractive summarization. 
\citet{gerani2014abstractive} used the parser of 
\citet{joty2013combining} to RST-parse product reviews. By extracting 
graph-based features, important aspects are identified in the review and 
included in the summary based on a template-based generation framework.  
Although the experiment shows that the RST can be beneficial for content 
selection, the proposed feature is rule-based and highly tailored to 
review documents. Instead, in this work, we extract a latent 
representation of the discourse directly from the 
\citet{yu2018transition} parser, and incorporate this into the 
abstractive summarizer.
	
	\subsection{Discourse Analysis for Document Classification}
    \citet{bhatia2015better} show that discourse analyses produced by an 
    RST parser can improve document-level sentiment analysis.  Based on 
 DPLP (Discourse Parsing from Linear Projection) ---
an RST parser by \citet{ji2014rep} --- they recursively propagate sentiment scores 
up to the root via a neural network. 
	
	A similar idea was proposed by \citet{lee2018adn}, where a recursive neural network is used to learn a discourse-aware representation. Here,  DPLP is utilized to obtain discourse structures, and a recursive neural network is applied to the doc2vec \cite{le2014distributed} representations for each EDU. The proposed approach is evaluated over sentiment analysis and sarcasm detection tasks, but found to not be competitive with benchmark methods. 
	
	Our work is different in that we use the latent representation (as distinct from the decoded discrete predictions) obtained from a neural RST parser. It is most closely related to the work of \citet{bhatia2015better} and \citet{lee2018adn}, but intuitively, our discourse representations contain richer information, and we evaluate over more tasks such as popularity prediction of online petitions.

    \section{Discourse Feature Extraction}

To incorporate discourse information into our models (for 
summarization or document regression), we use the RST parser developed by 
\newcite{yu2018transition} to extract shallow and latent discourse 
features.  The parser is competitive with other 
traditional parsers that use heuristic features 
\cite{feng2012text,li2014rec,ji2014rep} and other neural network-based 
parsers \cite{li2016dis}.

    \subsection{Shallow Discourse Features}

    Given a discourse tree produced by the RST parser 
    \cite{yu2018transition}, we compute several shallow features for an 
EDU: (1) the nuclearity score; (2) the relation score for each relation; 
and (3) the node type and that of its sibling. 

Intuitively, the nuclearity score measures how informative an EDU 
is, by calculating the (relative) number of ancestor nodes that are 
nuclei:\footnote{The ancestor nodes of an EDU are all the nodes 
traversed in its path to the root.}
\begin{equation*}
\frac{\sum_{x \in \text{ancestor}(e)} 
\mathbbm{1}_{\text{nucleus}}(x)}{h(\text{root})}
\end{equation*}
where $e$ is an EDU; $h(\text{x})$ gives the height from node 
$x$;\footnote{Note that tree height is computed from the leaves, and so 
the \emph{height} of the root node is equivalent to the \emph{depth} of 
a leave node.} and $\mathbbm{1}_{\text{nucleus}}(x)$ is an indicator 
function, i.e.\ it returns 1 when node $x$ is a nucleus and 0 
otherwise.

The relation score measures the importance of a discourse relation to an 
EDU, by computing the (relative) number of ancestor nodes that 
participate in the relation:
\begin{equation*}
\frac{\sum_{x \in \text{ancestor(e)}} \mathbbm{1}_{r_j}(x) h(x)}{\sum_{x 
\in \text{ancestor(e)}} h(x)}
\end{equation*}
where $r_j$ is a discourse relation (one of 18 in total).

Note that we weigh each ancestor node here by its height; our 
rationale is that ancestor nodes that are closer to the root are more 
important. The formulation of these shallow features (nuclearity and 
relation scores) are inspired by \newcite{ono1994abstract}, who propose 
a number of ways to score an EDU based on the RST tree structure. 
	

\begin{figure*}
    \centering
    \includegraphics[width=6in]{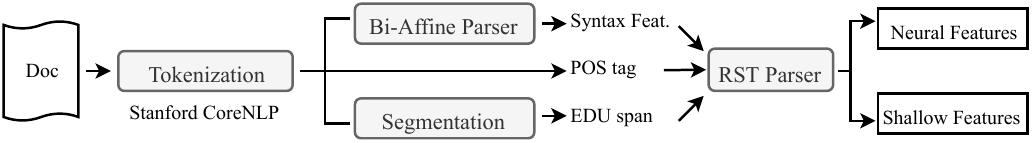}
    \caption{Pipeline of RST feature extraction}
    \label{fig:pipeline}
\end{figure*}

	
Lastly, we have 2 more features for the node type ({nucleus} or 
{satellite}) of the EDU and its sibling. In sum, our shallow feature representation
for an EDU has 21 dimensions: 1 nuclearity score, 18 relation scores, 
and 2 node types.
	
    \subsection{Latent Discourse Features}

In addition to the shallow features, we also extract latent features 
from the RST parser.

In the RST parser, each word and POS tag of an EDU span is first mapped 
to an embedding and concatenated to form the input sequence $\{x_1^w, 
..., x_m^w\}$ ($m$ is number of words in the EDU).   
\newcite{yu2018transition} also use syntax features ($\{x_1^s, ..., 
x_m^s\}$) from the bi-affine dependency parser  \cite{dozat2017deep}.  
The syntax features are the output of the multi-layer perceptron layer 
(see \newcite{dozat2017deep} for full details).
	

	
The two sequences are then fed to two (separate) bi-directional LSTMs 
and average pooling is applied to learn the latent representation for an 
EDU:
\begin{align*}
    \{h_1^w, .., h_m^w\} &= \text{Bi-LSTM}_1(\{x_1^w,.., x_m^w\}) \\
    \{h_1^s, ..., h_m^s\} &= \text{Bi-LSTM}_2(\{x_1^s,.., x_m^s\}) \\
    h^e &= \text{Avg-Pool}(\{h_1^w, .., h_m^w\}) \oplus \\
    &\phantom{= } \text{Avg-Pool}(\{h_1^s, ..., h_m^s\})
\end{align*}
where $\oplus$ denotes the concatenate operation.

Lastly, \newcite{yu2018transition} apply another bi-directional LSTM 
over the EDUs to learn a contextualized representation:
\begin{equation*}
    \{f_1, ...., f_n\} = \text{Bi-LSTM}(\{h_1^e,.., h_n^e\})
\end{equation*}

We extract the contextualized EDU representations ($\{f_1, ...., f_n\}$) 
    and use them as latent discourse features.


    \subsection{Feature Extraction Pipeline}

In \figref{pipeline}, we present the feature extraction pipeline.  Given 
an input document, we use Stanford CoreNLP to tokenize words and 
sentences, and obtain the POS 
tags.\footnote{\url{https://stanfordnlp.github.io/CoreNLP/}} We then 
parse the processed input with the bi-affine parser \cite{dozat2017deep} 
to get the syntax features.

The RST parser \cite{yu2018transition} requires EDU span information 
as input.  Previous studies have generally assumed the input text has been 
pre-processed to obtain EDUs, as state-of-the-art EDU 
segmentation models are very close to human performance
\cite{hernault2010hilda,ji2014rep}.  For our experiments, we use the 
pre-trained EDU segmentation model of \newcite{ji2014rep} to segment the 
input text to produce the EDUs.

Given the syntax features (from the bi-affine parser), POS tags, EDU 
spans, and tokenized text, we feed them to the RST parser to extract the 
shallow and latent discourse features.

	
We re-implemented the RST Parser in \textit{PyTorch} and were able to 
reproduce the results reported in the original paper.
We train the parser on the same data (385 documents from the Wall Street 
Journal), based on the configuration recommended in the paper.

To generate syntax features, we re-train an open-source bi-affine model,
and achieve over 95\% unlabelled and labelled attachment 
scores.\footnote{\url{https://github.com/XuezheMax/NeuroNLP2}} Source 
code used in our experiments is available at: \url{https://github.com/fajri91/RSTExtractor}.

\section{Abstractive Summarization}

\begin{figure}
    \centering
    \includegraphics[width=2.8in]{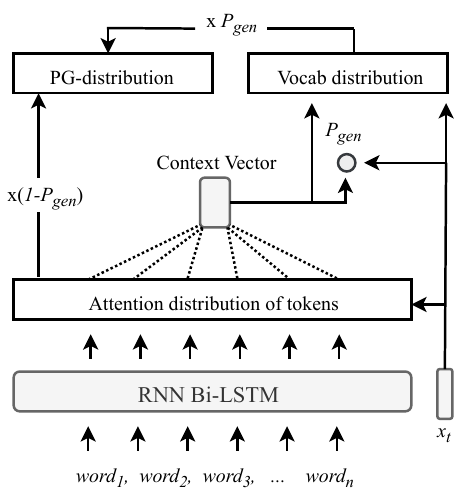}
    \caption{Architecture of the pointer--generator network 
    \cite{see2017get}.}
    \label{fig:summarization}
\end{figure}
	
Abstractive summarization is the task of creating a concise version of a 
document that encapsulates its core content.  Unlike extractive 
summarization, abstractive summarization has the ability to create new 
sentences that are not in the original document;  it is closer to how 
humans summarize, in that it generates paraphrases and blends multiple 
sentences in a coherent manner.
	
Current sequence-to-sequence models for abstractive summarization work 
like neural machine translation models, in largely eschewing symbolic analysis and 
learning purely from training data.  Pioneering work such as 
\newcite{rush2015neural}, for instance, assumes the neural architecture 
is able to learn main sentence identification, discourse structure 
analysis, and paraphrasing all in one model. Studies such as 
\newcite{gehrmann2018bottom,hsu2018unified} attempt to incorporate 
additional supervision (e.g.\ content selection) to improve 
summarization. Although there are proposals that extend 
sequence-to-sequence models based on discourse structure --- e.g.\ 
\newcite{Cohan+:2018} include an additional attention layer for 
document sections --- direct incorporation of discourse information is 
rarely  explored.
	
\newcite{hare1984direct} observe four core activities involved in 
creating a summary: (1) topic sentence identification; (2) deletion of 
unnecessary details; (3) paragraph collapsing; and (4) paraphrasing and 
insertion of connecting words. Current approaches  
\cite{nallapati2016abstractive,see2017get} capture topic sentence 
identification by leveraging the pointer network to do content 
selection, but the model is left to largely figure out the rest by 
providing it with a large training set, in the form of document--summary pairs. Our study 
attempts to complement the black-box model by providing additional 
supervision signal related to the discourse structure of a document.

\subsection{Summarization Model}
\label{sec:summ-model}

Our summarization model is based on the pointer--generator network 
\cite{see2017get}. We present the architecture in \figref{summarization},
and summarize it as follows:
\begin{align}
\{h_i\} &= \text{Bi-LSTM}_1(\{w_i\}) \label{eqn:first}\\
e_i^t&=v^{\intercal} \tanh(W_hh_i + W_ss_t + b_{e}) \label{eqn:attention}\\
a^t&=\softmax(e^t) \nonumber \\
h_t^*&=\sum_{i}a_i^th_i \nonumber \\
P_{voc}&= \softmax(V'(V[s_t,h_t^*]+b_{v})+b'_{v}) \nonumber \\
p_{gen}&=\sigma(w_{h^*}^{\intercal}h_t^*+w_s^{\intercal}s_t+w_x^{\intercal}x_t+b_{g}) 
\nonumber
\end{align}
where $\{h_i\}$ are the encoder hidden states, $\{w_i\}$ are the 
embedded encoder input words, $s_t$ is the decoder hidden state, and 
$x_t$ is the embedded decoder input word.

The pointer--generator network allows the model to either draw a word 
from its vocabulary (generator mode), or select a word from the input 
document (pointer mode). $p_{gen}$ is a scalar denoting the probability 
of triggering the generator mode, and $P_{voc}$ gives us the generator 
mode's vocabulary probability distribution.  To get the final 
probability distribution over all words, we sum up the attention weights 
and $P_{voc}$:
\begin{equation*}
P(w) = p_{gen}P_{voc}(w) + (1-p_{gen})\sum_{i:w_i=w}a_i^t 
\end{equation*}

To discourage repetitive summaries, \newcite{see2017get} propose adding 
coverage loss in addition to the cross-entropy loss:
\begin{align}
c^t &= \sum_{t'=0}^{t-1} a^{t'} \nonumber \\
e_i^t &= v^{\intercal} \text{tanh}(W_hh_i+W_ss_t+W_cc_i^t+b_{e}) 
\nonumber \\
\text{covloss}_t &=\sum_i \text{min} (a_i^t,c_i^t) \nonumber
\end{align}
Intuitively, the coverage loss works by first summing the attention 
weights over all words from previous decoding steps ($c^t$), using that 
information as part of the attention computation ($e_i^t$), and then 
penalising the model if previously attended words receive attention 
again (covloss$_t$). \newcite{see2017get} train the model for an 
additional 3K steps with the coverage penalty after it is trained with 
cross-entropy loss.

\begin{table*}[t]
    \begin{center}
        \begin{tabular}{lc@{\;\;}c@{\;\;}cc@{\;\;}c@{\;\;}cc@{\;\;}c@{\;\;}c}
\toprule
            \multicolumn{1}{c}{\multirow{2}{*}{\textbf{Method}}} & 
                \multicolumn{3}{c}{\textbf{F1}}                                                        
& \multicolumn{3}{c}{\textbf{Recall}}                                                    
& \multicolumn{3}{c}{\textbf{Precision}}                                                 
\\ \cline{2-10} \multicolumn{1}{c}{}                        & 
\multicolumn{1}{c}{R-1} & \multicolumn{1}{c}{R-2} & 
\multicolumn{1}{c}{R-L} & \multicolumn{1}{c}{R-1} & 
\multicolumn{1}{c}{R-2} & \multicolumn{1}{c}{R-L} & 
\multicolumn{1}{c}{R-1} & \multicolumn{1}{c}{R-2} & 
\multicolumn{1}{c}{R-L} \\
\midrule
		PG & 36.82 & 15.92 & 33.57 & 37.36 & 16.10 & 34.05 & 38.72 & 16.86 & 35.32 \\
		$+$M1-latent  & 37.76 & 16.51 & 34.48 & \textbf{40.15} & \textbf{17.52}\textbf{} & 36.65 & 37.90 & 16.64 & 34.61 \\
		$+$M1-shallow  & 37.45 & 16.23 & 34.22 & \textbf{40.15} & 17.38 & \textbf{36.68} & 37.34 & 16.24 & 34.13 \\
		$+$M2-latent  & \textbf{38.04} & \textbf{16.73} & \textbf{34.83} & 38.92 & 17.05 & 35.62 & \textbf{39.54} & \textbf{17.51} & \textbf{36.23} \\
		$+$M2-shallow & 37.15 & 16.13 & 33.96 & 38.52 & 16.68 & 35.21 & 38.19 & 16.67 & 34.91 \\
		$+$M3-latent  & 37.04 & 16.05 & 33.86 & 37.52 & 16.22 & 34.29 & 38.95 & 16.98 & 35.63 \\
		$+$M3-shallow  & 37.09 & 16.15 & 33.95 & 39.05 & 16.97 & 35.73 & 37.62 & 16.46 & 34.45 \\
\midrule
		PG$+$Cov & 39.32 & 17.22 & 36.02 & 40.33 & 17.61 & 36.93 & \textbf{40.82} & \textbf{17.99} & \textbf{37.42} \\ 
		$+$M1-latent & \textbf{40.06} & \textbf{17.63} & 36.70 & \textbf{44.44} & \textbf{19.53} & \textbf{40.69} & 38.60 & 17.05 & 35.39 \\
		$+$M1-shallow & 39.78 & 17.50 & 36.50 & 43.50 & 19.08 & 39.89 & 38.94 & 17.22 & 35.75 \\
		$+$M2-latent & 40.00 & 17.62 & \textbf{36.72} & 43.53 & 19.17 & 39.94 & 39.28 & 17.37 & 36.09 \\
		$+$M2-shallow & 39.58 & 17.30 & 36.36 & 44.00 & 19.19 & 40.38 & 38.40 & 16.87 & 35.31 \\
		$+$M3-latent & 39.23 & 17.00 & 36.00 & 42.95 & 18.54 & 39.37 & 38.29 & 16.69 & 35.16 \\
		$+$M3-shallow & 39.57 & 17.31 & 36.28 & 43.85 & 19.14 & 40.17 & 38.37 & 168.6 & 35.20 \\

\bottomrule
				
        \end{tabular}
    \end{center}
    \caption{\label{tab:font-table} Abstractive summarization results. }
\end{table*}

\subsection{Incorporating the Discourse Features}
\label{sec:incorporation}

We experiment with several simple methods to incorporate the discourse 
features into our summarization model. Recall that the discourse features 
(shallow or latent) are generated for each EDU, but the summarization 
model operates at the word level. To incorporate the features, we assume 
each word within an EDU span receives the same discourse feature representation.  Henceforth we use $g$ and $f$ to denote shallow and latent 
discourse features.

	
\textbf{Method-1 (M1)}: Incorporate the discourse features in the 
Bi-LSTM layer (\eqnref{first}) by concatenating them with the word 
embeddings:
\begin{align*}
\{h_i\} &= \text{Bi-LSTM}_1( \{w_i \oplus f_i\}); \text{or} \\
\{h_i\} &= \text{Bi-LSTM}_1( \{w_i \oplus g_i\})
\end{align*}
	
\textbf{Method-2 (M2)}: Incorporate the discourse features by adding 
another Bi-LSTM:
\begin{align*}
    \{h_i'\} &= \text{Bi-LSTM}_2(\{h_i \oplus f_i\}); \text{or} \\
    \{h_i'\} &= \text{Bi-LSTM}_2(\{h_i \oplus g_i\})
\end{align*}
	
\textbf{Method-3 (M3)}: Incorporate the discourse features in the 
    attention layer (\eqnref{attention}):
\begin{align*}
e_i^t &= v^\intercal \tanh(W_hh_i + W_ss_t + W_{f}f_i+ b_{e}); \text{or} 
\\
e_i^t &= v^\intercal \tanh(W_hh_i + W_ss_t + W_{g}g_i+ b_{e})
\end{align*}
	
\subsection{Data and Result}

We conduct our summarization experiments using the anonymized CNN/DailyMail corpus 
\cite{nallapati2016abstractive}. We follow the data preprocessing steps 
in \newcite{see2017get} to obtain 287K training examples, 13K 
validation examples, and 11K test examples. 
	
All of our experiments use the default hyper-parameter configuration of 
\newcite{see2017get}. Every document and its summary pair are truncated 
to 
400 and 100 tokens respectively (shorter texts are padded accordingly).  
    The model has 256-dimensional hidden states and 128-dimensional word 
embeddings, and vocabulary is limited to the most frequent 50K tokens.  
During test inference, we similarly limit the length of the input 
document to 400 words and the length of the generated summary to 35--100 
words for beam search.
	
Our experiment has two pointer--generator network baselines: (1) one without 
the coverage mechanism (``PG''); and (2) one with the coverage mechanism 
(``PG$+$Cov''; \secref{summ-model}). For each baseline, we incorporate 
the latent and shallow discourse features separately in 3 ways 
(\secref{incorporation}), giving us 6 additional results.

We train the models for approximately 240,000-270,000 iterations (13 
epochs). When we include the coverage mechanism (second baseline), we train 
for an additional 3,000--3,500 iterations using the coverage penalty, 
following  \newcite{see2017get}. 
	
We use ROUGE \cite{lin2004rouge} as our evaluation metric, which is a 
standard measure based on overlapping n-grams between the generated 
summary and the reference summary.  We assess unigram (R-1), bigram 
(R-2), and longest-common-subsequence (R-L) overlap, and present F1, 
recall and precision scores in \tabref{font-table}.
	
For the first baseline (PG), we see that incorporating discourse 
features consistently improves recall and F1. This observation is 
consistent irrespective of how (e.g.\ M1 or M2) and what (e.g.\ shallow 
or latent features) we add. These improvements do come at the expense of 
precision, with the exception of M2-latent (which produces small 
improvements in precision). Ultimately however, the latent features are in 
general a little better, with M2-latent produing the best results based 
on F1.

	\begin{figure}[t]
		\centering
        \includegraphics[width=\linewidth]{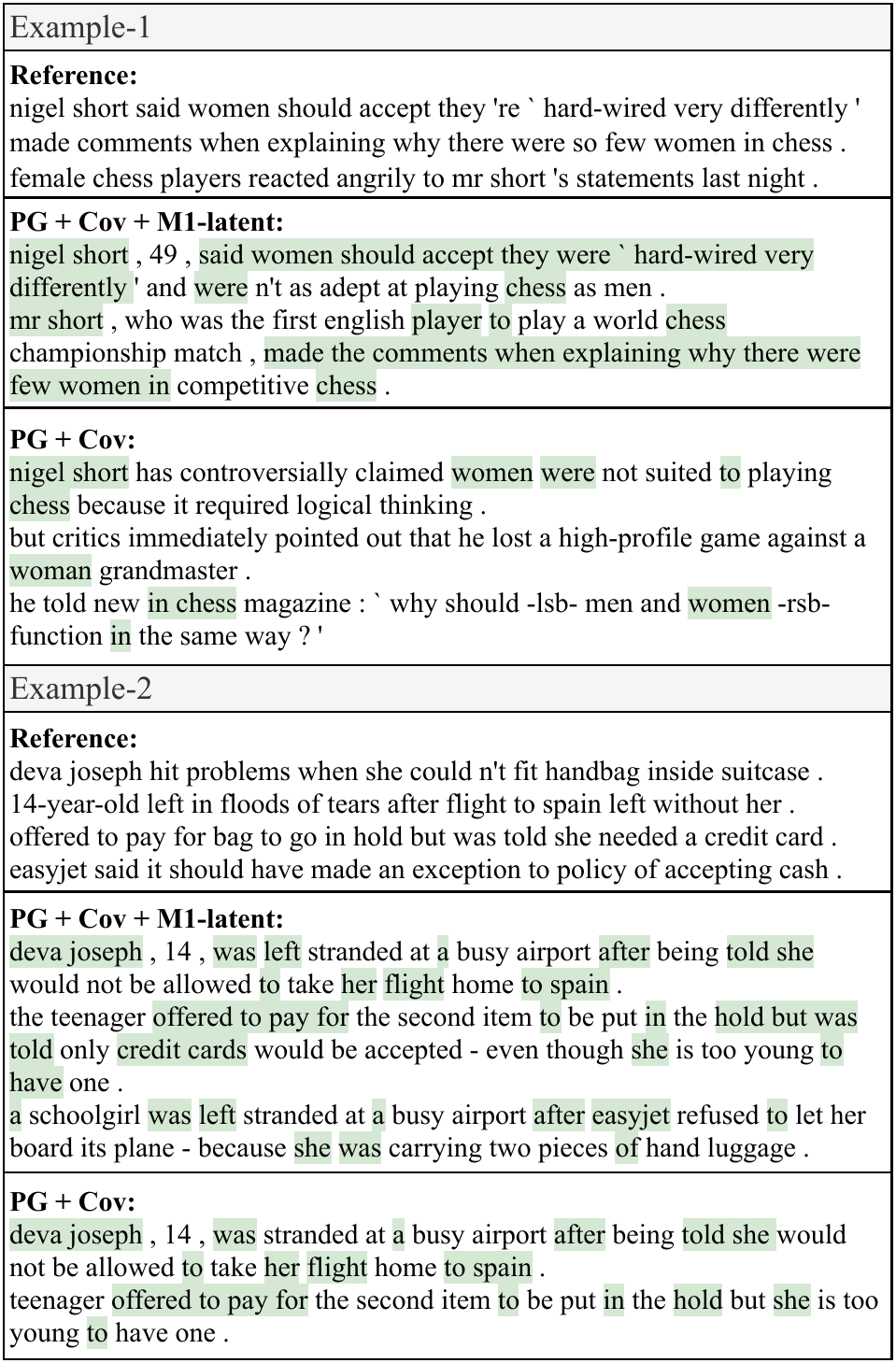}
        \caption{Comparison of summaries between our model and the 
        baseline.}
		\label{fig:summary}
	\end{figure}
	
We see similar observations for the second baseline (PG$+$Cov):  
recall is generally improved at the expense of precision. In terms of 
F1, the gap between the baseline and our models is a little closer, and 
M1-latent and M2-latent are the two best performers. 

	
	
	\subsection{Analysis and Discussion}
	
	
We saw previously that our models generally improve recall.  To better 
understand this, we present 2 examples of generated summaries, one by 
the baseline (``PG$+$Cov'') and another by our model (``M1-latent''), in 
\figref{summary}.  The highlighted words are overlapping words in the 
reference.  In the first example, we notice that the summary generated 
by our model is closer to the reference, while the baseline has other 
unimportant details (e.g.\ \textit{he told new in chess magazine : ` why 
should -lsb- men and women -rsb- function in the same way ?}).  In the 
second example, although there are more overlapping words in our model's 
summary, it is a little repetitive (e.g.\ first and third paragraph) and 
less concise.

Observing that our model generally has better recall 
(\tabref{font-table}) and its summaries tend to be more verbose (e.g.\ 
second example in \figref{summary}), we calculated the average length of 
generated summaries for PG$+$Cov and M1-latent, and found that they are 
of length 55.2 and 64.4 words respectively. This suggests that although 
discourse information helps the summarization model overall (based on 
consistent improvement in F1), the negative side effect is that the 
summaries tend to be longer and potentially more repetitive.

    \section{Petition Popularity Prediction}
	
Online petitions are open letters to policy-makers or governments
 requesting change or action, based on the support of members of society at large. Understanding the factors that determine the 
popularity of a petition, i.e.\ the number of supporting signatures it 
will receive, provides valuable information for institutions or 
independent groups to communicate their goals  
\cite{proskurnia2017predicting}.
	
\newcite{subramanian2018content} attempt to model petition popularity
by utilizing the petition text.  One novel contribution is that 
they incorporate an auxiliary ordinal regression objective that predicts 
the scale of signatures (e.g.\ 10K vs.\ 100K). Their results demonstrate 
that the incorporation of auxiliary loss and hand-engineered features 
boost performance over the baseline.

In terms of evaluation metric, \newcite{subramanian2018content} use: (1)  
mean absolute error (MAE); and (2) mean absolute percentage error 
(MAPE), calculated as
$\frac{100}{n} \sum_{i=1}^n \frac{\hat{y}_i - y_i}{y_i}$, where $n$ is 
the number of examples and $\hat{y}_i$ ($y_i$) the predicted (true) 
value. Note that in both cases, lower numbers are better.
	
Similar to the abstractive summarization task, we experiment with 
incorporating the discourse features of the petition text to the 
petition regression model, under the hypothesis that discourse structure should benefit the model.
	
\subsection{Deep Regression Model}

As before, our model is based on the model of  
\newcite{subramanian2018content}. The input is a concatenation of the 
petition's title and content words, and the output is the log number of 
signatures. The input sequence is mapped to GloVe vectors 
\cite{pennington2014glove} and processed by several convolution filters 
with max-pooling to create a fixed-width hidden representation, which is 
then fed to fully connected layers and ultimately activated by an 
exponential linear unit to predict the output.  The model is optimized 
with mean squared error (MSE). In addition to the MSE loss, the authors 
include an auxiliary ordinal regression objective that predicts the 
scale of signatures (e.g.\ $\{10, 100, 1000, 10000, 100000\}$), and 
found that it improves performance. Our model is based on the best model 
that utilizes both the MSE and ordinal regression loss.


	
	
    \subsection{Incorporating the Discourse Features}

We once again use the methods of incorporation presented in 
\secref{incorporation}.  As the classification model uses convolution 
networks, only Method-1 is directly applicable.

We also explore replacing the convolution networks with a bidirectional 
LSTM (``Bi-LSTM w/ GloVe''), based on the idea that recurrent networks 
are better at capturing long range dependencies between words and EDUs.  
For this model, we test both Method-1 and Method-2 to 
incorporate the discourse features.\footnote{Our LSTM has 
200 hidden units, and uses a dropout rate of 0.3, and L2 regularization.}

Lastly, unlike the summarization model that needs word level input  (as 
the pointer network requires words to attend to in the source document),  
we experiment with replacing the input words with EDUs, and embed the 
EDUs with either the latent (``Bi-LSTM w/ latent'') or the shallow 
(``Bi-LSTM w/ shallow'') features.

	
	\subsection{Data, Result, and Discussion}
	
We use the US Petition dataset from 
\cite{subramanian2018content}.\footnote{The data is collected from 
\url{https://petitions.whitehouse.gov}.}  In total we have 1K petitions 
with over 12M signatures after removing petitions that have less than 
150 signatures. We use the same train/dev/test split of 80/10/10 as 
    \newcite{subramanian2018content}.
	
\begin{table}[t]
    \begin{center}
    \begin{adjustbox}{max width=\linewidth}
        \begin{tabular}{lcc}
            \toprule
            \textbf{Model} & \textbf{MAE} & \textbf{MAPE} \\
            \midrule
            CNN w/ GloVe & 1.16 & 14.38 \\
            $+$ M1-latent & 1.15 & 14.66 \\
            $+$ M1-shallow & \textbf{1.12}\anoteone & \textbf{14.19} \\
            \midrule
            Bi-LSTM w/ GloVe & 1.14 & 14.57 \\
            $+$ M1-latent & 1.13 & 14.39 \\
            $+$ M1-shallow & 1.13 & 14.25 \\
            $+$ M2-latent & 1.12 & 14.02 \\
            $+$ M2-shallow & 1.13 & 14.20\\
            \midrule
            Bi-LSTM w/ latent & \textbf{1.11}\anotetwo & \textbf{13.91} \\
            Bi-LSTM w/ shallow & 1.15 & 14.67 \\
            \bottomrule
        \end{tabular}
    \end{adjustbox}
    \end{center}
    \caption{\label{tab:petition} Average petition regression
      performance over 3 runs (noting that lower is better for both MAE
      and MAPE). One-sided t-tests show that both (1) and
      (2) are significantly better than the baseline ($p < 0.05$ and
      $p<0.005$, resp.). }
\end{table}


We present the test results in \tabref{petition}. We tune the 
models based on the development set using MAE, and find that most  
converge after 8K--10K iterations of training. We are able to reproduce 
the performance of the baseline model (``CNN w/ GloVe''), and find that
once again, adding the shallow discourse features improves results.

Next we look at the LSTM model (``Bi-LSTM w/ GloVe''). Interestingly, we
found that replacing the CNN with an LSTM results in improved MAE, but
worse MAPE. Adding discourse features to this model generally has marginal improvement 
in all cases.

When we replace the word sequence with EDUs (``Bi-LSTM w/ latent'' and 
``Bi-LSTM w/ shallow''), we see that the latent features outperform the 
shallow features. This is perhaps unsurprising, given that the shallow 
discourse features have no information about the actual content, and are  
unlikely to be effective when used in isolation without the word features.

	\section{Conclusion and Future Work}
	
In this paper, we explore incorporating discourse information into two 
tasks: abstractive summarization and  popularity prediction of online petitions. We 
experiment with both hand-engineered shallow features and latent 
features extracted from a neural discourse parser, and found that adding 
them generally benefits both tasks. The caveat, however, is that the 
best method of incorporation and feature type (shallow or latent) 
appears to be task-dependent, and so it remains to be seen whether we can 
find a robust universal approach for incorporating discourse information 
into NLP tasks.
	
	
	\bibliography{aaai20}
    \bibliographystyle{acl_natbib}
	
\end{document}